\documentclass[11pt]{article}
\PassOptionsToPackage{table}{xcolor}
\usepackage[preprint]{acl}
\usepackage{times}
\usepackage{latexsym}
\usepackage[T1]{fontenc}
\usepackage[utf8]{inputenc}
\usepackage{microtype}
\usepackage{inconsolata}
\usepackage{graphicx}
\usepackage{booktabs}
\definecolor{gainbg}{RGB}{225,242,229}
\usepackage{pifont}

\newcommand{\xmark}{\ding{55}}
\usepackage{amsmath,amssymb,amsthm}
\usepackage{algorithm}
\usepackage{algorithmic}
\usepackage{multirow}
\usepackage{enumitem}
\usepackage{subcaption}
\usepackage{array}
\usepackage{xcolor}
\usepackage{tabularx}
\usepackage{siunitx}
\usepackage{hyperref,url,overpic}
\newcommand{\harnessmdp}{\textsc{HarnessMDP}}
\definecolor{originalfiguregreen}{RGB}{33,121,64}

\newtheorem{theorem}{Theorem}
\newtheorem{proposition}{Proposition}
\newtheorem{definition}{Definition}

\newtheorem{corollary}{Corollary}

\title{\harnessmdp{}: Learning to Control LLM Agent Harnesses with Offline Reinforcement Learning}
\author{
  Haiwen Yi\thanks{Equal contribution.} \\
  University of Toronto \\
  Canada
  \And
  Xinyuan Song\footnotemark[1]\thanks{Corresponding author.} \\
  Emory University \\
  United States
}

\begin{document}
\hypersetup{pdfauthor={Anonymous},pdftitle={Learning to Control LLM Agent Harnesses with Offline Reinforcement Learning}}
\maketitle

\begin{abstract}
The execution harness is a critical component of LLM agents, shaping how the agent interacts with its environment and organizes multi-step execution. Prior work has shown that improving tools, workflows, and other harness components can substantially improve agent performance. This suggests that the harness can be treated as a learnable control layer that adapts execution decisions from experience, rather than as a predetermined execution configuration. In this work, we introduce \harnessmdp{}, a learning-based framework that makes harness control itself learnable. \harnessmdp{} formulates agent execution as a finite-horizon Markov decision process, where the harness learns to select structural execution actions from the evolving agent state. The harness policy is learned offline from collected trajectories using task performance and deployed online to adaptively control new executions. Extensive experiments on coding tasks, $\tau$-bench, AgentBench, and other agent evaluation settings show that \harnessmdp{} consistently improves execution behavior and final task performance over baseline harnesses. These results demonstrate the effectiveness of \harnessmdp{} and establish learnable harness control as a practical mechanism for improving LLM agents through better execution. Our code is available at \url{https://github.com/Hik289/Agentic-RL-harness.git}.
\end{abstract}

\vspace{-1em}
\section{Introduction}
\vspace{-0.5em}

\begin{figure*}[!ht]
\centering
\includegraphics[width=0.8\textwidth]{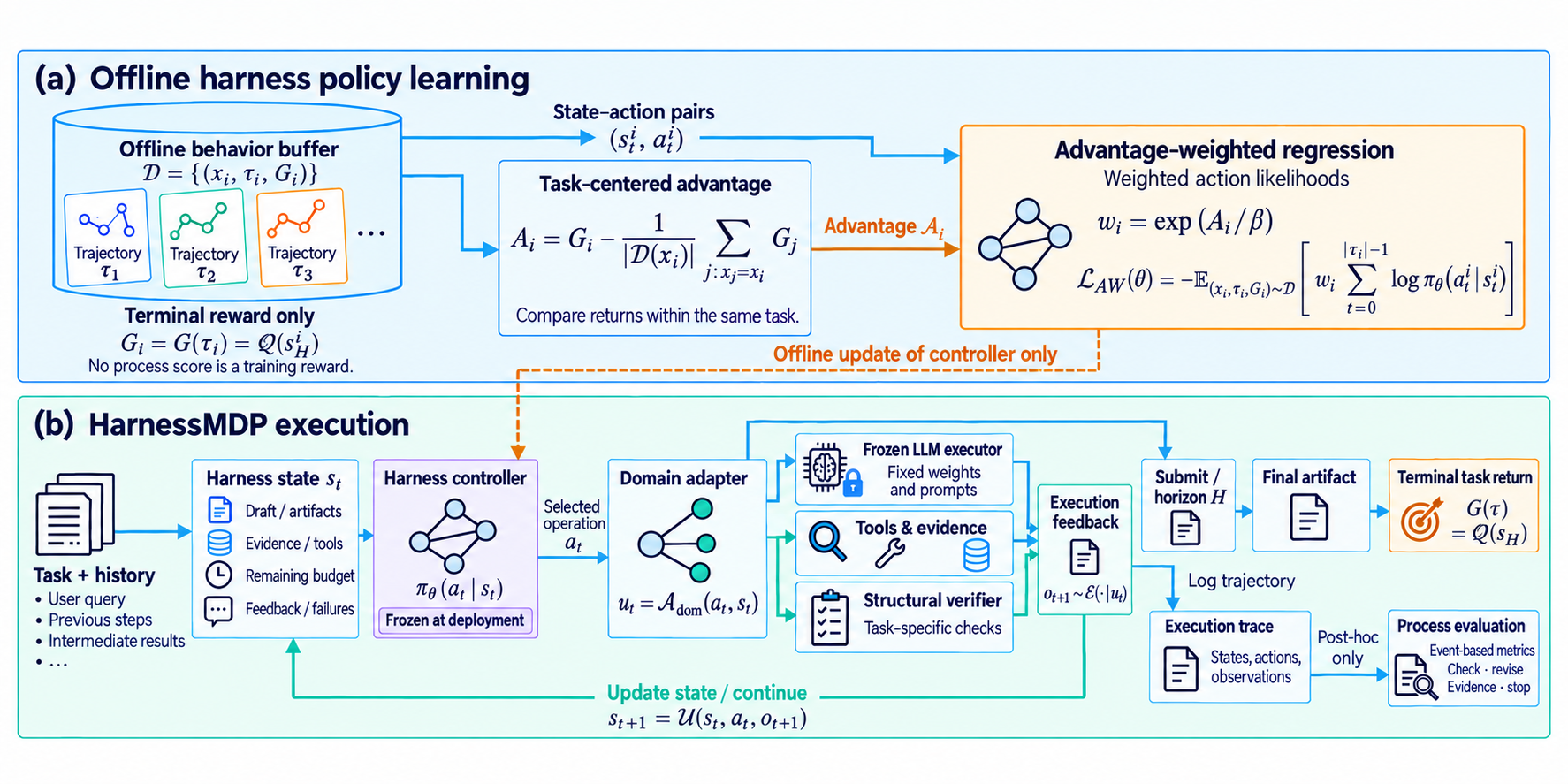}
\vspace{-1.5em}
\caption{Overview of \harnessmdp{}. \textbf{(1)} Collect offline trajectories and task returns. \textbf{(2)} Learn the harness controller with advantage-weighted regression. \textbf{(3)} Deploy the controller for state-dependent execution through domain adapters. \textbf{(4)} Evaluate task outcomes and process behavior separately, using only task return $G$ for training.}
\label{fig:intuition}
\vspace{-1em}
\end{figure*}

Large language model (LLM) agents increasingly solve complex tasks through multi-step interaction with external environments \citep{react2023,agentbench2023,gaia2023}. Their performance depends not only on the underlying model, but also on the execution harness that determines how the model interacts with the environment. Tool interfaces can substantially affect agent behavior \citep{toolformer2023,sweagent2024}, while workflows and agent scaffolds can improve coordination across multiple steps \citep{autogen2023,metagpt2023,agentverse2023}. The harness is particularly important in interactive benchmarks such as SWE-bench \citep{swebench2024}, $\tau$-bench \citep{taubench2024}, and ToolSandbox \citep{toolsandbox2024}, where success depends on a sequence of decisions and actions rather than a single model response. Recent evidence further shows that different harnesses can lead the same underlying model to substantially different behaviors \citep{harnessbench2026}. These findings suggest that improving the harness is an important way to improve LLM agents.

Existing work has improved agent harnesses by designing better tools \citep{toolformer2023}, interfaces \citep{sweagent2024}, workflows \citep{autogen2023}, and multi-agent scaffolds \citep{metagpt2023,agentverse2023}. However, these designs typically specify in advance how the agent should execute a task. We instead study whether an agent can learn which harness operation to use at each step. Depending on what has happened so far, the harness may choose to gather more information, verify an earlier result, revise a previous step, or continue execution. We therefore formulate harness control as a sequential decision problem and learn a policy that selects harness operations based on the current state of the agent trajectory.

We introduce \harnessmdp{}, a framework that formulates harness control as a finite-horizon Markov decision process. As illustrated in Figure~\ref{fig:intuition}, \harnessmdp{} has four stages: \textbf{(1)} collect offline execution trajectories $\tau=(s_1,a_1,\ldots,s_T,a_T)$ with task return $G(\tau)$, providing the support from which the policy can learn \citep{levine2020offline,kumar2020conservative}; \textbf{(2)} learn a harness policy $\pi$ that maximizes expected task return,
\begin{equation}
\pi^\star=\arg\max_{\pi}\mathbb{E}_{\tau\sim p_\pi(\tau)}\!\left[G(\tau)\right],
\end{equation}
where $p_\pi(\tau)$ is the trajectory distribution induced by $\pi$; \textbf{(3)} deploy the learned policy online to select operations from \emph{observe}, \emph{retrieve}, \emph{call-tool}, \emph{draft}, \emph{check}, \emph{revise}, and \emph{submit} according to the evolving execution state $s_t$; and \textbf{(4)} evaluate both final task outcomes and execution behavior. Domain adapters map selected operations to executable actions while the underlying LLM remains frozen.

We extensively evaluate \harnessmdp{} on coding tasks, adapted $\tau$-bench Retail \citep{taubench2024}, adapted AgentBench DB-Bench \citep{agentbench2023}, and controlled agent environments covering knowledge work, research, multi-tool interaction, long-memory tasks, and planning. \harnessmdp{} consistently improves execution behavior and improves final task performance over baseline harnesses in multiple settings. In particular, final quality increases by $18.2$ percentage points on adapted $\tau$-bench Retail, $13.2$ points on adapted AgentBench DB-Bench, and $10.0$ points on coding under the calibrated structural verifier. These results demonstrate that harness control can be learned effectively from offline agent experience and used to improve subsequent execution.

Our contributions are summarized as follows:
\begin{itemize}[leftmargin=*,nosep]
\item We introduce \harnessmdp{}, which formulates the execution harness as a finite-horizon MDP and learns state-dependent control over structural agent operations.
\item We develop an offline learning mechanism based on task-centered advantage-weighted regression \citep{awr2019}, learning harness policies from task performance and deploying them for adaptive online execution.
\item We distinguish task outcomes from post-hoc process behavior and characterize how finite offline support determines which execution patterns and outcome improvements can be learned.
\item We extensively evaluate \harnessmdp{} on coding, $\tau$-bench \citep{taubench2024}, AgentBench \citep{agentbench2023}, and controlled agent environments, demonstrating improved execution behavior and final task performance.
\end{itemize}

% Related Work is retained verbatim below for source traceability and typeset
% from sections/related_work.tex in the appendix.

\begin{table*}[!ht]
\centering
\caption{Comparison with closely related methods. \checkmark{} indicates that the method explicitly supports the property.}
\vspace{-0.5em}
\label{tab:related_work_novelty}
\setlength{\tabcolsep}{6pt}
\resizebox{\textwidth}{!}{
\begin{tabular}{@{}lccccc@{}}
\toprule
\textbf{Method}
& \textbf{Stepwise Control}
& \textbf{Harness Actions}
& \textbf{Offline Learning}
& \textbf{Frozen LLM}
& \textbf{Outcome-Only Learning} \\
\midrule
AFlow \citep{aflow2025}
& \xmark & \xmark & \xmark & \checkmark & \checkmark \\
ScoreFlow \citep{scoreflow2025}
& \xmark & \xmark & \checkmark & \xmark & \checkmark \\
GEPA \citep{gepa2025}
& \xmark & \xmark & \xmark & \checkmark & \checkmark \\
RHO \citep{rho2026}
& \xmark & \checkmark & \xmark & \checkmark & \xmark \\
Retrospex \citep{retrospex2024}
& \checkmark & \xmark & \checkmark & \checkmark & \xmark \\
GLIDER \citep{glider2025}
& \checkmark & \xmark & \checkmark & \xmark & \checkmark \\
Harness Control \citep{paul2026harness}
& \checkmark & \checkmark & \xmark & \checkmark & \checkmark \\
MetaCtrl \citep{metactrl2026}
& \checkmark & \checkmark & \xmark & \checkmark & \xmark \\
ReasonRAG \citep{processvsoutcome2025}
& \checkmark & \xmark & \checkmark & \xmark & \xmark \\
\midrule
\rowcolor{black!8}
\textbf{\harnessmdp{} (Ours)}
& \checkmark
& \checkmark
& \checkmark
& \checkmark
& \checkmark \\
\bottomrule
\end{tabular}
}
\vspace{-1.5em}
\end{table*}

\vspace{-0.5em}
\section{Related Work}

\vspace{-0.5em}
\paragraph{Agent and tool-use benchmarks.} Interactive agent benchmarks evaluate increasingly long-horizon execution: AgentBench \citep{agentbench2023} covers diverse interactive environments, $\tau$-bench \citep{taubench2024} targets tool-agent-user interaction, GAIA \citep{gaia2023} evaluates real-world assistant tasks, and WebArena \citep{webarena2024} and WebShop \citep{webshop2022} study web interaction. SWE-bench \citep{swebench2024} evaluates repository-level coding, while AgentGym \citep{agentgym2024}, ToolBench \citep{toolllm2023}, and ToolSandbox \citep{toolsandbox2024} emphasize interactive and tool-mediated execution; long-memory benchmarks \citep{locolmo2024} further test information use over extended contexts. Rather than improving the underlying executor on these tasks, we study whether the external policy controlling its execution can itself be learned.
\vspace{-0.5em}
\paragraph{Prompt, workflow, and harness optimization.} Automatic prompt optimization searches over instructions or demonstrations \citep{ape2023,opro2024}, while DSPy \citep{dspy2023} and TextGrad \citep{textgrad2024} optimize compound language-model programs. More closely related, AFlow \citep{aflow2025}, ScoreFlow \citep{scoreflow2025}, and AutoPDL \citep{autopdl2025} search over agent workflows, while GEPA \citep{gepa2025} uses execution feedback to optimize prompts. Recent harness-evolution approaches \citep{ahe2026,rho2026,thelastharness2026} directly modify the surrounding agent program, instructions, skills, or tools. In contrast, \harnessmdp{} keeps these execution components fixed and learns a state-conditioned policy that chooses among harness operations during each trajectory.
\vspace{-0.5em}
\paragraph{Reflection, planning, and agent control.} Reflexion \citep{reflexion2023} uses verbal feedback for iterative improvement, Voyager \citep{voyager2023} accumulates reusable skills, LATS \citep{lats2024} searches over reasoning and action trajectories, and ADaPT \citep{adapt2024} adaptively decomposes difficult tasks. Interactive agent learning in WebShop \citep{webshop2022} and AgentGym \citep{agentgym2024} instead trains policies through environment interaction. These approaches motivate adaptive execution, but \harnessmdp{} learns only the external structural controller: the LLM executor remains frozen, and the learned actions are harness operations rather than solution tokens or environment actions.

% \paragraph{Offline policy learning.} Our learning setting is closely related to offline RL, where policies are learned from previously collected experience without additional interaction \citep{offline_rl_review2020,brandfonbrener2021}. We instantiate harness learning with advantage-weighted regression (AWR) \citep{awr2019}, which reweights observed actions according to their estimated advantages and is related to advantage-weighted actor-critic \citep{awac2020} and implicit Q-learning \citep{iql2021}. Behavior cloning \citep{pomerleau1989,ross2011imitation} provides the corresponding unweighted imitation baseline, while Forced CHECK tests whether improvements can be reproduced by mechanically adding verification rather than learning state-dependent control.
\vspace{-0.5em}
\paragraph{Process evaluation and reward design.} Process-level evaluation studies behavior that final task scores alone may not reveal, including intermediate reasoning and execution quality \citep{processvsoutcome2025,encodabench2025,strainedcoherence2026}. We similarly analyze verification, evidence use, revision, and stopping, but keep these measurements separate from the learning reward. This distinction follows classical reward-shaping results \citep{ng1999policy}: arbitrary auxiliary rewards can alter the task-optimal policy. We therefore train \harnessmdp{} only from task performance and use process measurements solely for post-hoc analysis.
\vspace{-0.5em}
\paragraph{Positioning and novelty.} As summarized in Table~\ref{tab:related_work_novelty}, \harnessmdp{} combines stepwise harness control with offline policy learning while keeping the executor fixed, and separates task supervision from process evaluation.

\vspace{-0.5em}
\section{Method}\label{sec:method}
\vspace{-0.5em}

We present \harnessmdp{}, which learns an external control policy around a frozen LLM executor. Rather than modifying model parameters, prompts, or generated solution tokens, \harnessmdp{} learns when to invoke structural execution operations from previously collected trajectories. We first formulate harness execution as a finite-horizon MDP and then describe how its policy is learned offline from task outcomes.
\vspace{-0.5em}
\subsection{Harness Control as an MDP}

We model harness execution as a finite-horizon MDP $\mathcal{M}=(\mathcal{S},\mathcal{A},P,G,H)$. At step $t$, the harness observes state $s_t$ and selects a structural operation
\begin{equation}
a_t\sim\pi_\theta(\cdot\mid s_t).
\label{eq:harness-policy}
\end{equation}
A domain adapter executes the operation and updates the state, while the LLM executor remains frozen.
\vspace{-0.5em}
\paragraph{State.} The harness state contains observable information relevant to execution control, including trajectory progress, intermediate artifacts, evidence, tool outputs, verifier feedback, failures, remaining budget, and previous operations. It excludes hidden LLM activations, so the controller conditions only on the observable execution process.
\vspace{-0.5em}
\paragraph{Action space.} We use the shared structural action space $\mathcal{A}$=\{\textsc{observe},\textsc{retrieve},\textsc{call-tool},\textsc{draft},\textsc{check},\textsc{revise},\textsc{submit}\}. These actions control \emph{how} execution proceeds rather than \emph{what} the LLM generates. Domain adapters instantiate each operation and mask invalid actions, separating the shared control policy from domain-specific execution.

\vspace{-0.5em}
\subsection{Task Objective}

An execution produces a trajectory $\tau=(s_0,a_0,s_1,a_1,\ldots,s_H)$ with task quality $G(\tau)=\mathcal{Q}(s_H)$, where $\mathcal{Q}$ is a domain-specific task rubric. We assign no intermediate reward to particular harness operations, i.e., $r_t=0$ for $t<H-1$. The controller is trained solely to maximize task performance:
\begin{equation}
\pi^\star=\arg\max_{\pi}\mathbb{E}_{\tau\sim p_\pi(\tau)}\left[G(\tau)\right].
\label{eq:harness-objective}
\end{equation}
Thus, operations such as \textsc{check} and \textsc{revise} receive no intrinsic reward; they become more likely only when they occur in executions with better task outcomes.
\vspace{-0.5em}
\subsection{Offline Harness Policy Learning}

We learn the controller from a fixed offline behavior buffer $\mathcal{D}=\{(x_i,\tau_i,G_i)\}_{i=1}^{N}$, where $x_i$ identifies the task and $G_i=G(\tau_i)$ is the terminal return. No additional environment interaction is required during policy optimization.

Because task difficulty varies, we compare executions within the same task rather than using raw returns directly. For each trajectory, we define the task-centered advantage and its corresponding weight as
\begin{equation}
\small
A_i=G_i-\frac{1}{|\mathcal{D}(x_i)|}\sum_{j:x_j=x_i}G_j, \quad w_i=\exp(A_i/\beta),
\label{eq:trajectory-advantage}
\end{equation}
where $\mathcal{D}(x_i)$ contains the trajectories collected for task $x_i$ and $\beta>0$ controls the strength of reweighting. Since supervision is available only at the trajectory level, all decisions in $\tau_i$ share the same advantage, $A(s_t^i,a_t^i)=A_i$.

We optimize the harness controller by exponentially reweighting the action likelihoods of each trajectory according to its task-centered advantage:
\begin{equation}
\small
\mathcal{L}_{\mathrm{AW}}(\theta)=-\mathbb{E}_{(x_i,\tau_i,G_i)\sim\mathcal{D}}\left[w_i\sum_{t=0}^{|\tau_i|-1}\log\pi_\theta(a_t^i\mid s_t^i)\right].
\label{eq:awr-objective}
\end{equation}
This objective gives greater weight to decisions from executions that outperform other trajectories for the same task and less weight to those from lower-performing executions. Because supervision is trajectory-level, the same weight is applied to every decision within a trajectory.

Because learning is restricted to a fixed behavior buffer, \harnessmdp{} can amplify useful execution patterns already represented in the data but cannot directly learn strategies outside its support \citep{offline_rl_review2020,brandfonbrener2021}. Offline trajectory diversity is therefore an explicit constraint on what the learned harness can recover.
\vspace{-0.5em}
\subsection{Deployment}

After training, both the harness controller $\pi_\theta$ and the LLM executor are frozen. Starting from $s_0$, the controller selects a harness operation, the domain adapter converts it into an executable action, and the resulting observation updates the harness state:
\begin{equation}
\begin{aligned}
a_t &\sim \pi_\theta(\cdot\mid s_t), \qquad u_t=\mathcal{A}_{\mathrm{dom}}(a_t,s_t),\\
o_{t+1} &\sim \mathcal{E}(\cdot\mid u_t), \qquad s_{t+1}=\mathcal{U}(s_t,a_t,o_{t+1}).
\end{aligned}
\label{eq:deployment-loop}
\end{equation}
Here, $\mathcal{A}_{\mathrm{dom}}$ maps shared harness operations to domain-specific actions, allowing the same controller to operate across different environments. Execution continues until \textsc{submit} or the horizon $H$ is reached. Thus, \harnessmdp{} controls the execution procedure while the underlying task environment and executor remain unchanged.

\vspace{-0.5em}
\section{Theory}
\label{sec:theory}
\vspace{-0.5em}
We analyze whether reward design preserves task-optimal policies and how advantage-weighted reweighting affects execution behavior relative to terminal outcomes.

\vspace{-0.5em}
\subsection{Task-Consistent Reward Design}
\label{sec:reward-theory}

We use finite-horizon potential-based reward shaping~\citep{ng1999policy} to characterize task-consistent rewards. Our experiments use $\Phi_t\equiv 0$, assigning rewards only through terminal task quality.

\begin{definition}[Task-rubric return]
\label{def:task-rubric-return}
Let $G(\tau)=\mathcal Q(s_H)$ denote the terminal return in Section~\ref{sec:method}. The task objective is
\begin{equation}
J_G(\pi)=\mathbb E_{\tau\sim p_\pi}[G(\tau)],
\label{eq:task-rubric-objective}
\end{equation}
where $p_\pi$ is induced by the initial-state distribution, harness policy, and frozen-LLM transitions.
\end{definition}

\begin{definition}[Potential-shaped task reward]
\label{def:potential-shaped-reward}
Let $\{\Phi_t:\mathcal S\rightarrow\mathbb R\}_{t=0}^{H}$ be bounded potentials with $\Phi_H=0$. Define
\begin{equation}
\small
r_t^\Phi(s_t,a_t,s_{t+1})=r_t(s_t,a_t,s_{t+1})+\Phi_{t+1}(s_{t+1})-\Phi_t(s_t),
\label{eq:potential-shaped-reward}
\end{equation}
where $r_t=0$ for $t<H-1$ and $r_{H-1}=G(\tau)$. The shaped objective is
\begin{equation}
J_\Phi(\pi)=\mathbb E_{\tau\sim p_\pi}\left[\sum_{t=0}^{H-1}r_t^\Phi(s_t,a_t,s_{t+1})\right].
\label{eq:potential-shaped-objective}
\end{equation}
\end{definition}

\begin{theorem}[Optimal-policy invariance]
\label{thm:optimal-policy-invariance}
Under Definitions~\ref{def:task-rubric-return} and~\ref{def:potential-shaped-reward},
\begin{equation}
J_\Phi(\pi)=J_G(\pi)-C_{\rho_0},
\label{eq:objective-equivalence}
\end{equation}
where $C_{\rho_0}=\mathbb E_{s_0\sim\rho_0}[\Phi_0(s_0)]$ is independent of $\pi$. Consequently,
\begin{equation}
\operatorname*{argmax}_{\pi}J_\Phi(\pi)=\operatorname*{argmax}_{\pi}J_G(\pi).
\label{eq:optimal-policy-set}
\end{equation}
\end{theorem}

Theorem~\ref{thm:optimal-policy-invariance} shows that potential shaping preserves task-optimal policies. Thus, intermediate feedback can be introduced without changing the objective. Our implementation uses terminal rewards without explicit action bonuses. See Appendix~\ref{app:proof-optimal-policy-invariance}.

\begin{proposition}[Action-pattern rewards can change the optimum]
\label{prop:action-pattern-counterexample}
There exists a finite-horizon Harness MDP and an action-pattern bonus $b(a)$ such that
\begin{equation}
\operatorname*{argmax}_{\pi}\mathbb E_\pi[G(\tau)]\neq\operatorname*{argmax}_{\pi}\mathbb E_\pi[G(\tau)+b(a_0)].
\label{eq:action-pattern-noninvariance}
\end{equation}
\end{proposition}

Proposition~\ref{prop:action-pattern-counterexample} shows that directly rewarding actions such as \textsc{check} or \textsc{revise} may favor unnecessary steps. This motivates learning their utility from task outcomes rather than prescribing action bonuses. See Appendix~\ref{app:proof-action-pattern-counterexample}.

\vspace{-0.5em}
\subsection{Offline Reweighting and Process--Outcome Decoupling}
\label{sec:finite-buffer-theory}

We characterize process changes under idealized trajectory reweighting induced by the task-centered advantages in Equation~\eqref{eq:trajectory-advantage}. This analysis concerns the collected trajectories rather than the exact deployment distribution.

\begin{definition}[Advantage-weighted trajectory distribution]
\label{def:finite-buffer-aw}
Let $\mathcal D=\{\tau^{(1)},\ldots,\tau^{(N)}\}$ be the collected trajectories with empirical distribution $\mu_{\mathcal D}$. For task-centered advantage $A(\tau)$ and temperature $\beta>0$, define
\begin{equation}
q_{\mathrm{AW}}(\tau)=\frac{\mu_{\mathcal D}(\tau)\exp(A(\tau)/\beta)}{\mathbb E_{\mu_{\mathcal D}}[\exp(A/\beta)]}.
\label{eq:aw-trajectory-distribution}
\end{equation}
\end{definition}

Writing $w(\tau)=\exp(A(\tau)/\beta)$, this distribution assigns greater weight to higher-advantage trajectories, consistent with Equation~\eqref{eq:awr-objective}.

\begin{theorem}[Process change under advantage weighting]
\label{thm:finite-buffer-decoupling}
For any bounded process statistic $\Psi:\mathcal D\rightarrow\mathbb R$, define $\Delta\Psi=\mathbb E_{q_{\mathrm{AW}}}[\Psi]-\mathbb E_{\mu_{\mathcal D}}[\Psi]$. Then
\begin{equation}
\Delta\Psi=\frac{\operatorname{Cov}_{\mu_{\mathcal D}}(w,\Psi)}{\mathbb E_{\mu_{\mathcal D}}[w]}.
\label{eq:exact-process-shift}
\end{equation}
Thus, $\Delta\Psi$ is nonnegative precisely when $\operatorname{Cov}_{\mu_{\mathcal D}}(w,\Psi)$ is nonnegative.
\end{theorem}

Theorem~\ref{thm:finite-buffer-decoupling} shows that process changes depend on their covariance with trajectory weights, not on explicit rewards for particular actions. See Appendix~\ref{app:proof-finite-buffer-decoupling}.

\begin{corollary}[Process improvement under monotone advantage weighting]
\label{cor:monotone-process-improvement}
If a measurable non-decreasing function $h:\mathbb R\rightarrow\mathbb R$ satisfies
\begin{equation}
\mathbb E_{\mu_{\mathcal D}}[\Psi(\tau)\mid A(\tau)]=h(A(\tau)),
\end{equation}
then $\operatorname{Cov}_{\mu_{\mathcal D}}(w,\Psi)\geq 0$ and $\Delta\Psi\geq 0$.
\end{corollary}

Corollary~\ref{cor:monotone-process-improvement} shows that behaviors more prevalent in higher-advantage trajectories gain weight without action-level rewards. See Appendix~\ref{app:proof-monotone-process-improvement}.

\begin{proposition}[Terminal-return correlation does not determine process change]
\label{prop:return-correlation-not-decisive}
There exist offline trajectory distributions, returns $G$, process statistics $\Psi$, and task-centered advantages $A$ such that
\begin{equation}
\operatorname{Cov}_{\mu_{\mathcal D}}(G,\Psi)<0,\qquad \operatorname{Cov}_{\mu_{\mathcal D}}(w,\Psi)>0.
\label{eq:opposite-process-covariances}
\end{equation}
Consequently, $\Delta\Psi>0$ despite a negative marginal correlation between terminal return and process behavior.
\end{proposition}

Proposition~\ref{prop:return-correlation-not-decisive} shows that task-centered weighting can favor behaviors associated with stronger within-task performance even when their marginal return correlation is negative. See Appendix~\ref{app:proof-return-correlation-not-decisive}.

Together, these results explain why process and outcome changes need not coincide. They characterize idealized trajectory reweighting rather than the deployed state-dependent controller, whose behavior is evaluated empirically.

\vspace{-0.5em}
\section{Experiments}\label{sec:experiments}
\vspace{-0.5em}

\subsection{Experimental Setup}
\label{sec:experimental-setup}

\paragraph{Tasks and protocol.}
We evaluate \harnessmdp{} across six controlled domains: knowledge-work~\citep{assistantbench2024}, coding, research QA, multi-tool, long-memory~\citep{longmemeval2025}, and planning~\citep{planbench2023}. Each domain contains 100 human-annotated tasks (20 easy, 60 standard, and 20 hard), with 80 used for offline training and 20 held out for evaluation. The stratified splits use disjoint task identifiers, templates, entities, distractors, and starter artifacts to reduce leakage. We additionally evaluate two public-benchmark adapters, $\tau$-bench retail~\citep{taubench2024} and AgentBench DB-Bench~\citep{agentbench2023}, each with 16 training and 20 held-out tasks. These adapters preserve the original tool-interaction or database-reasoning structure but use our task-rubric evaluator rather than official benchmark scoring. All policies are evaluated with three random seeds and three rollouts per held-out task, yielding 180 evaluation episodes per setting.
\vspace{-0.5em}
\paragraph{Accuracy and process evaluation.}
We report \emph{Accuracy} as the mean normalized terminal rubric score across evaluation episodes. For a trajectory $\tau$, the score is
\begin{equation}
\mathrm{Acc}(\tau)=\frac{\sum_{c\in\mathcal C(x)}\mathrm{score}_c(\tau)}{\sum_{c\in\mathcal C(x)}\mathrm{max\_score}_c},
\label{eq:accuracy-metric}
\end{equation}
where $\mathcal C(x)$ denotes the task-specific scoring criteria. Knowledge-work evaluates correctness, evidence support, coverage, and format~\citep{assistantbench2024}; coding evaluates unit-test correctness, parseability, safety, and cost~\citep{swebench2024}; research QA evaluates correctness, evidence support, citations, and format~\citep{hotpotqa2018}; multi-tool evaluates correctness and tool validity~\citep{taubench2024,agentbench2023}; long-memory evaluates correctness and source-session matching~\citep{longmemeval2025,locolmo2024}; and planning evaluates constraint satisfaction and plan quality~\citep{planbench2023}. Coding uses \textbf{executable unit tests} for its main correctness criterion~\citep{swebench2024}, whereas the remaining domains combine \textbf{structural verification and rubric-based scoring} against human-annotated criteria. The scorer receives the task, criterion, and candidate output without access to the reference answer. Importantly, \textbf{the same scoring procedure} supplies terminal training rewards and evaluation scores, while calibration is restricted to verifier validation. Beyond final Accuracy, we separately measure execution behavior through \textbf{event-based process metrics}, including verification, revision after failure, evidence use, and stopping when sufficient, motivated by process-level agent evaluation and feedback-based refinement~\citep{react2023,reflexion2023,madaan2023selfrefine}. These detectors use fixed heuristic thresholds and may miss semantically valid behaviors.
\vspace{-0.5em}
\paragraph{Training and baselines.}
We train the harness controller on trajectories collected from the Base and exploratory harnesses using the task-centered advantage-weighted objective in Equation~\eqref{eq:awr-objective}, with weights $\exp(A/\beta)$~\citep{awr2019}. We compare the learned controller (\emph{Learned}) against the original harness (\emph{Base}), behavior cloning (\emph{Imitation})~\citep{pomerleau1989,ross2011imitation}, and mandatory verification (\emph{Forced CHECK}). Imitation tests whether reproducing observed execution patterns is sufficient, while Forced CHECK tests whether requiring verification alone explains the improvements. The underlying LLM executor remains frozen across methods.
\vspace{-0.5em}
\subsection{Learned CHECK Behavior}

We next test whether training changes the controller's use of the \textsc{check} action. For each evaluation trajectory, we record whether at least one \textsc{check} occurs before the final \textsc{submit}, and report the fraction of trajectories satisfying this condition. Table~\ref{tab:check-rate} shows that this event is rare under the Base policy but becomes substantially more frequent under the Learned policy across all eight settings.
\begin{equation}
\small
\mathrm{CheckRate}=\frac{1}{N}\sum_{i=1}^{N}\mathbf{1}\left[\exists\,t<t_{\mathrm{sub}}^{(i)}:\,a_t^{(i)}=\textsc{check}\right].
\label{eq:check-rate}
\end{equation}

\begin{table}[!ht]
\vspace{-1em}
\centering
\small
\caption{\textbf{CHECK-before-submit rate.} Percentage of evaluation trajectories containing at least one \textsc{check} action before the final \textsc{submit}. Base and Learned denote trajectories generated by the original and learned harness policies, respectively.}
\label{tab:check-rate}
\vspace{-1em}
\begin{tabular}{@{}lcc@{}}
\toprule
\textbf{Setting} & \textbf{Base (\%)} & \textbf{Learned (\%)} \\
\midrule
Knowledge-work & 0.0 & 5.6 \\
Coding & 0.0 & 17.8 \\
Research & 0.0 & 13.9 \\
Multi-tool & 0.0 & 8.9 \\
Long-memory & 0.0 & 10.0 \\
Planning & 0.6 & 6.7 \\
\midrule
$\tau$-bench retail & 0.0 & 17.2 \\
AgentBench DB-Bench & 0.0 & 16.7 \\
\bottomrule
\end{tabular}
\vspace{-1em}
\end{table}

\vspace{-0.5em}
\subsection{Final Task Quality}

\begin{figure}[!ht]
\vspace{-0.5em}
\centering
\includegraphics[width=\linewidth]{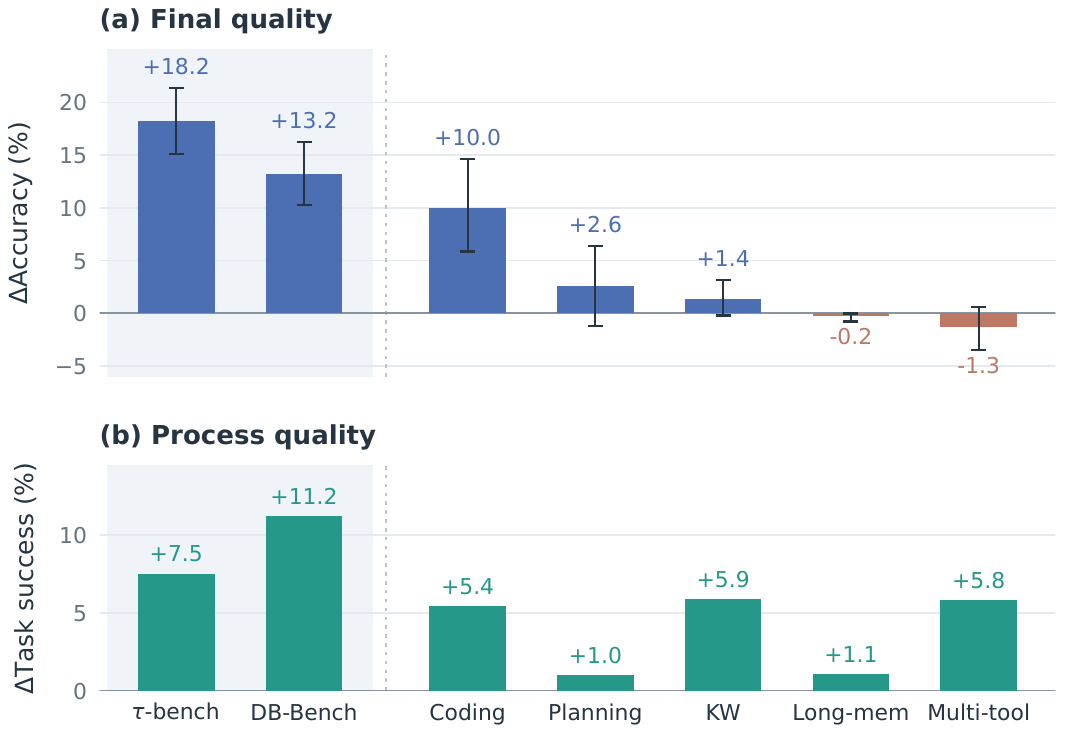}
\vspace{-1em}
\caption{\textbf{Accuracy and task-success changes under Learned.} The horizontal axis lists the eight evaluation settings. The vertical axis reports the change from Base to Learned in percentage terms: panel (a) shows accuracy change ($\Delta \mathrm{Acc}$), and panel (b) shows task-success change ($\Delta \mathrm{TS}$). Positive values indicate improvement.}
\label{fig:f5}
\vspace{-1em}
\end{figure}

We examine whether learned changes in execution behavior translate into better terminal outcomes. Figure~\ref{fig:f5} compares changes in accuracy ($\Delta \mathrm{Acc}$) and task success ($\Delta \mathrm{TS}$) across all eight settings. Task-success gains are broad, whereas the largest accuracy gains concentrate in $\tau$-bench retail, DB-Bench, and coding. Table~\ref{tab:benchmarks} reports these three settings in detail, with accuracy improvements of $18.2\%$, $13.2\%$, and $10.0\%$, respectively.

\begin{table}[!ht]
\vspace{-0.5em}
\centering
\footnotesize
\caption{\textbf{Largest accuracy gains.} Base Acc and Learned Acc are mean accuracy scores, and $\Delta \mathrm{Acc}$ is the improvement from Base to Learned, expressed as a percentage. The two benchmark adapters use our Harness MDP scoring protocols rather than official upstream scores.}
\label{tab:benchmarks}
\vspace{-1em}
\begin{tabular}{@{}lccc@{}}
\toprule
\textbf{Setting} & \textbf{Base Acc} & \textbf{Learned Acc} & \textbf{$\Delta \mathrm{Acc}$ (\%)} \\
\midrule
$\tau$-bench retail & 0.337 & 0.519 & $\mathbf{+18.2}$ \\
DB-Bench & 0.415 & 0.547 & $\mathbf{+13.2}$ \\
Coding & 0.712 & 0.812 & $\mathbf{+10.0}$ \\
\bottomrule
\end{tabular}
\vspace{-1em}
\end{table}

\vspace{-0.5em}
\subsection{Baseline ablations.}
We compare our advantage-weighted learned policy (Learned) with behavior cloning (Imitation) and mandatory verification before submission (Forced CHECK). As shown in Table~\ref{tab:bc-fc}, Learned outperforms Imitation in all eight settings and both baselines in five. These results suggest that the gains cannot be explained by imitation or additional verification alone. Instead, advantage-weighted learning enables state-dependent decisions about when checking and revision are beneficial, yielding particularly large accuracy gains in coding, $\tau$-bench retail, and DB-Bench.

\begin{table*}[!ht]
\vspace{-0.5em}
\centering
\small
\caption{\textbf{Baseline ablations.}
Accuracy changes (\%) relative to Base for Learned, Imitation, and Forced CHECK. The final two columns compare Learned against each baseline; green cells indicate positive advantages.}
\label{tab:bc-fc}
\vspace{-1em}
\resizebox{0.8\textwidth}{!}{%
\begin{tabular}{@{}lrrrrr@{}}
\toprule
\textbf{Setting}
& \textbf{Learned}
& \textbf{Imitation}
& \textbf{Forced CHECK}
& \textbf{Learned -- Imitation}
& \textbf{Learned -- Forced CHECK} \\
\midrule
Knowledge-work
& $+1.4$ & $-0.4$ & $+0.1$
& \cellcolor{gainbg}\textbf{$+1.8$}
& \cellcolor{gainbg}\textbf{$+1.3$} \\

Coding
& $+10.0$ & $-8.3$ & $+0.0$
& \cellcolor{gainbg}\textbf{$+18.3$}
& \cellcolor{gainbg}\textbf{$+10.0$} \\

Research
& $-0.3$ & $-4.2$ & $-0.2$
& \cellcolor{gainbg}\textbf{$+3.8$}
& $-0.1$ \\

Multi-tool
& $-1.3$ & $-6.8$ & $+0.3$
& \cellcolor{gainbg}\textbf{$+5.5$}
& $-1.6$ \\

Long-memory
& $-0.3$ & $-5.8$ & $+0.0$
& \cellcolor{gainbg}\textbf{$+5.5$}
& $-0.3$ \\

Planning
& $+2.6$ & $+0.0$ & $-2.3$
& \cellcolor{gainbg}\textbf{$+2.6$}
& \cellcolor{gainbg}\textbf{$+4.9$} \\

$\tau$-bench retail
& $+18.2$ & $+8.2$ & $+0.1$
& \cellcolor{gainbg}\textbf{$+10.0$}
& \cellcolor{gainbg}\textbf{$+18.1$} \\

DB-Bench
& $+13.2$ & $+5.8$ & $-0.5$
& \cellcolor{gainbg}\textbf{$+7.4$}
& \cellcolor{gainbg}\textbf{$+13.7$} \\

\bottomrule
\end{tabular}%
}
\vspace{-0.5em}
\end{table*}

\vspace{-0.5em}
\subsection{Results on Controlled Domains.}
Table~\ref{tab:main-synthetic} reports results across six domains. Coding achieves the largest accuracy gain (+10.0\%) under the calibrated structural verifier, which evaluates parseability, safety, cost compliance, testing, and checking. The original strict coding rubric was already near saturation, limiting further improvement. Planning and knowledge-work show smaller gains, while other domains exhibit negligible or slightly negative accuracy changes. Task success improves in five of six domains, particularly knowledge-work and multi-tool, whereas increased early submission offsets the checking benefits in research. Overall, learned harness control improves task success more consistently than accuracy, with an average task-success gain of +2.8\%.

\begin{table*}[!ht]
\vspace{-0.5em}
\centering
\small
\caption{\textbf{Results on controlled domains.}
Accuracy and task-success changes from Base to Learned across six domains. Each policy is evaluated on 20 held-out tasks with three seeds and three rollouts per task ($N=180$). CHECK Rate measures the percentage of Learned trajectories containing a CHECK action before submission. Green cells indicate positive improvements.}
\label{tab:main-synthetic}
\vspace{-1em}
\resizebox{0.8\textwidth}{!}{%
\begin{tabular}{@{}lrrrrr@{}}
\toprule
\textbf{Domain}
& \textbf{Base Acc}
& \textbf{Learned Acc}
& \textbf{$\Delta$Acc (\%)}
& \textbf{$\Delta$Task Success (\%)}
& \textbf{CHECK Rate (\%)} \\
\midrule
Knowledge-work
& 0.450 & 0.464
& \cellcolor{gainbg}$+1.4$
& \cellcolor{gainbg}$+5.9$
& 5.6 \\

\textbf{Coding}
& \textbf{0.712} & \textbf{0.812}
& \cellcolor{gainbg}$\mathbf{+10.0}$
& \cellcolor{gainbg}$+5.4$
& 17.8 \\

Multi-tool
& 0.528 & 0.515
& $-1.3$
& \cellcolor{gainbg}$+5.8$
& 8.9 \\

Planning
& 0.385 & 0.412
& \cellcolor{gainbg}$+2.6$
& \cellcolor{gainbg}$+1.0$
& 6.7 \\

Research
& 0.314 & 0.311
& $-0.3$
& $-2.6$
& 13.9 \\

Long-memory
& 0.453 & 0.451
& $-0.3$
& \cellcolor{gainbg}$+1.1$
& 10.0 \\

\midrule
\textit{Macro mean}
& -- & --
& \cellcolor{gainbg}\textit{$+2.0$}
& \cellcolor{gainbg}\textit{$+2.8$}
& \textit{10.5} \\
\bottomrule
\end{tabular}%
}
\vspace{-0.5em}
\end{table*}

\vspace{-0.5em}
\subsection{Process--Outcome Decoupling}
\label{sec:process-outcome-decoupling}

\begin{figure}[!ht]
\vspace{-0.5em}
\centering
\includegraphics[width=0.9\linewidth]{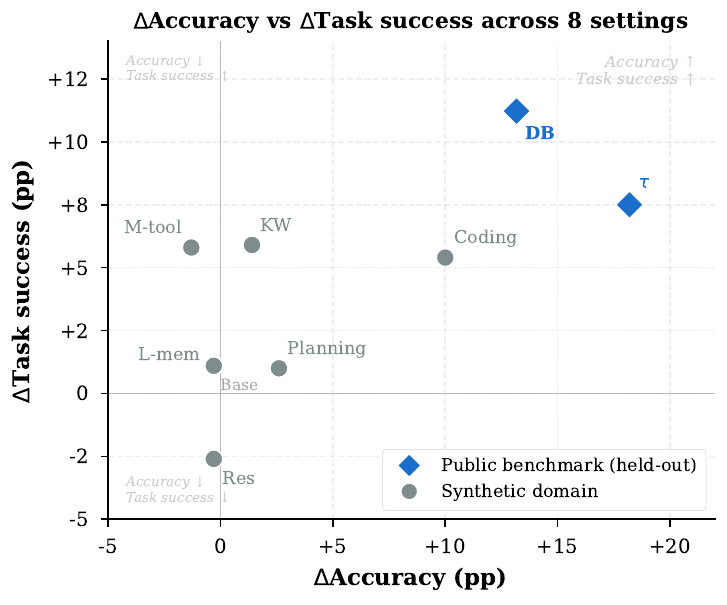}
\vspace{-1em}
\caption{\textbf{Process--outcome decoupling across evaluation settings.} Changes in Accuracy and Task Success from Base to Learned across six controlled domains and two public-benchmark adapters.}
\label{fig:dg-hms-scatter}
\vspace{-1em}
\end{figure}

Figure~\ref{fig:dg-hms-scatter} compares changes in execution behavior and final task quality across eight evaluation settings. Both Accuracy and Task Success improve on the public-benchmark adapters, whereas several controlled domains exhibit positive process gains despite negligible or negative accuracy changes. These observations are consistent with Theorem~\ref{thm:finite-buffer-decoupling} and Proposition~\ref{prop:return-correlation-not-decisive}, which characterize how advantage-weighted process changes can differ from marginal terminal-return associations. Although these results concern idealized offline reweighting rather than the deployed controller, the experiments reinforce the importance of evaluating process quality and terminal outcomes separately.

\vspace{-0.5em}
\subsection{Event-Level Behavioral Analysis.}
Table~\ref{tab:full-hms} examines how learned harness control changes execution behavior across six domains. The most consistent improvement is verification before submission, increasing from 1 to 113 of 1,080 episodes overall. Coding also shows more frequent testing and revision after failure. However, other behaviors remain largely unchanged, while premature submission increases substantially in research. These results indicate that learned control primarily strengthens verification rather than uniformly improving all execution behaviors.

\vspace{-0.5em}
\subsection{Sensitivity to Verifier Calibration}
\label{sec:coverage-calibration}

\begin{figure}[!ht]
\vspace{-0.5em}
\centering
\includegraphics[width=0.95\linewidth]{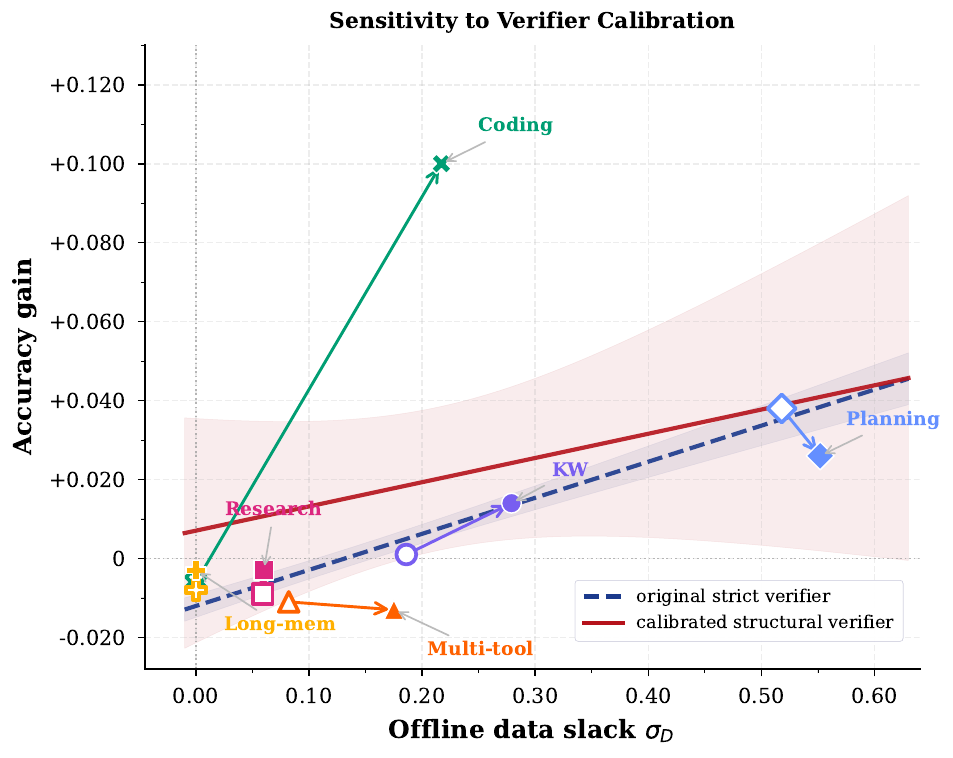}
\vspace{-1em}
\caption{Offline data slack versus accuracy gains under strict (hollow) and calibrated (filled) verifiers.}
\label{fig:sigma}
\vspace{-1em}
\end{figure}

Figure~\ref{fig:sigma} examines the relationship between offline data slack $\sigma_D$ and accuracy gains under two verifier configurations. The original strict verifier produces near-saturated coding scores and an apparent strong correlation. After calibration, coding exhibits moderate slack and the correlation weakens, indicating that offline data coverage does not consistently predict accuracy improvements.

\begin{table*}[!ht]
\vspace{-0.5em}
\centering
\small
\caption{\textbf{Changes in execution behavior.}
Event-rate changes (\%) from Base to Learned. Positive values indicate increased frequency. Green cells highlight favorable behavioral changes. N/A indicates that an event is not applicable to the corresponding domain.}
\label{tab:full-hms}
\vspace{-1em}
\resizebox{0.75\textwidth}{!}{%
\begin{tabular}{@{}lrrrrrr@{}}
\toprule
\textbf{Behavior}
& \textbf{Knowledge}
& \textbf{Coding}
& \textbf{Research}
& \textbf{Multi-tool}
& \textbf{Memory}
& \textbf{Planning} \\
\midrule
Check Before Submit
& \cellcolor{gainbg}+5.6
& \cellcolor{gainbg}+17.8
& \cellcolor{gainbg}+13.9
& \cellcolor{gainbg}+8.9
& \cellcolor{gainbg}+10.0
& \cellcolor{gainbg}+6.1 \\

Test Before Submit
& N/A
& \cellcolor{gainbg}+10.0
& N/A
& N/A
& N/A
& N/A \\

Revision After Failure
& 0.0
& \cellcolor{gainbg}+4.3
& 0.0
& 0.0
& 0.0
& \cellcolor{gainbg}+3.4 \\

Evidence Before Claim
& 0.0
& N/A
& 0.0
& 0.0
& -0.6
& N/A \\

Early Submit
& +0.5
& +0.6
& +25.0
& 0.0
& +3.3
& \cellcolor{gainbg}-3.3 \\

\bottomrule
\end{tabular}%
}
\vspace{-1em}
\end{table*}

\vspace{-0.5em}
\subsection{Sensitivity to Process-Event Thresholds}
\label{sec:threshold-sensitivity}

Table~\ref{tab:es_thresh} examines how the EarlySubmit threshold affects measured Task Success. Increasing the threshold substantially changes the research results: at $t=0.35$, all Base research episodes are classified as premature submissions, inflating the apparent improvement. We adopt $t=0.25$ to avoid detector saturation. This demonstrates that process measurements depend on detector calibration.

\begin{table}[!ht]
\vspace{-0.5em}
\centering
\small
\caption{Task Success changes (\%) under different EarlySubmit thresholds.}
\label{tab:es_thresh}
\vspace{-1em}
\resizebox{\linewidth}{!}{%
\begin{tabular}{@{}lccc@{}}
\toprule
\textbf{Domain / Measure}
& $t=0.25$
& $t=0.30$
& $t=0.35$ \\
\midrule
Knowledge-work & $+5.9$ & $+6.6$ & $+6.6$ \\
Coding & $+5.4$ & $+5.4$ & $+5.4$ \\
Research & $\mathbf{-2.6}$ & $+0.2$ & $+9.6$ \\
Multi-tool & $+5.8$ & $+5.8$ & $+5.8$ \\
Long-memory & $+1.1$ & $+0.2$ & $+0.2$ \\
Planning & $+1.0$ & $+0.7$ & $+0.7$ \\
\midrule
\textbf{Macro mean} & $\mathbf{+2.8}$ & $\mathbf{+3.2}$ & $\mathbf{+4.7}$ \\
Positive domains & 5/6 & 6/6 & 6/6 \\
Base Research EarlySubmit (\%) & 0.0 & 20.0 & 100.0 \\
\bottomrule
\end{tabular}%
}
\vspace{-1em}
\end{table}

\vspace{-0.5em}
\section{Conclusion}
\vspace{-0.5em}
We introduce \harnessmdp{}, a framework for learning execution control around frozen LLM agents through offline reinforcement learning. By optimizing structural harness actions, the learned controller improves verification behavior and achieves notable accuracy gains across several domains and benchmarks. Our findings highlight the potential of learned harness control while revealing the importance of offline data quality and the distinction between execution behavior and task outcomes. Future work will explore broader environments, online adaptation, and more robust evaluation protocols.

\section*{Limitations}
This work has several limitations. First, our evaluation covers a finite set of tasks and environments, which may not capture the complexity and variability of real-world agent deployments. Second, the effectiveness of the learned controller depends on the quality, diversity, and coverage of offline training data, potentially limiting generalization to unseen scenarios. Third, evaluation metrics and verification procedures may introduce biases that affect the interpretation of experimental results. Finally, our framework focuses on controlling a frozen LLM rather than improving its underlying capabilities. Future work could explore broader environments, more diverse models, improved evaluation protocols, and adaptive learning from online interactions.

\bibliography{references}

\clearpage
\appendix
% Appendix uses the normal caption and float separation, without manual vspace.

\section{Proofs and Mathematical Background}
\label{app:theory}

\subsection{Proof of Theorem~\ref{thm:optimal-policy-invariance}}
\label{app:proof-optimal-policy-invariance}

By Definition~\ref{def:potential-shaped-reward}, the cumulative shaped return of any trajectory $\tau=(s_0,a_0,\ldots,s_H)$ satisfies
\begin{equation}
\begin{aligned}
&\sum_{t=0}^{H-1}r_t^{\Phi}(s_t,a_t,s_{t+1})\\
&=\sum_{t=0}^{H-1}\left[r_t(s_t,a_t,s_{t+1})+\Phi_{t+1}(s_{t+1})-\Phi_t(s_t)\right]\\
&=\sum_{t=0}^{H-1}r_t(s_t,a_t,s_{t+1})+\Phi_H(s_H)-\Phi_0(s_0)\\
&=G(\tau)-\Phi_0(s_0),
\end{aligned}
\label{eq:proof-shaped-sum}
\end{equation}
where the second equality follows from telescoping, and the last uses $\Phi_H(s_H)=0$ and the terminal-reward definition in Definition~\ref{def:potential-shaped-reward}.

Taking expectations over the trajectory distribution induced by $\pi$ yields
\begin{equation}
\begin{aligned}
J_{\Phi}(\pi)
&=\mathbb{E}_{\tau\sim p_\pi}\left[\sum_{t=0}^{H-1}r_t^{\Phi}(s_t,a_t,s_{t+1})\right]\\
&=\mathbb{E}_{\tau\sim p_\pi}[G(\tau)]-\mathbb{E}_{s_0\sim\rho_0}[\Phi_0(s_0)]\\
&=J_G(\pi)-C_{\rho_0},
\end{aligned}
\label{eq:proof-policy-equivalence}
\end{equation}
where $C_{\rho_0}=\mathbb{E}_{s_0\sim\rho_0}[\Phi_0(s_0)]$ is independent of $\pi$.

Since $C_{\rho_0}$ is independent of the policy, the constant shift preserves the ordering of all policies under the two objectives. Consequently,
\begin{equation}
\operatorname*{argmax}_{\pi}J_{\Phi}(\pi)=\operatorname*{argmax}_{\pi}J_G(\pi).
\label{eq:proof-argmax-equivalence}
\end{equation}
This establishes Equation~\eqref{eq:optimal-policy-set} and completes the proof of Theorem~\ref{thm:optimal-policy-invariance}.

\subsection{Proof of Proposition~\ref{prop:action-pattern-counterexample}}
\label{app:proof-action-pattern-counterexample}

Consider a one-step Harness MDP with initial state $s_0$, action space $\mathcal A=\{\textsc{submit},\textsc{check}\}$, and terminal states $s_{\mathrm{good}}$ and $s_{\mathrm{bad}}$. The action \textsc{submit} deterministically reaches $s_{\mathrm{good}}$, whereas \textsc{check} reaches $s_{\mathrm{bad}}$. Let $G(s_{\mathrm{good}})=1$ and $G(s_{\mathrm{bad}})=0$.

Under the task-rubric objective in Equation~\eqref{eq:task-rubric-objective}, the deterministic policies satisfy
\begin{equation}
\begin{aligned}
J_G(\pi_{\mathrm{submit}})&=\mathbb E_{\pi_{\mathrm{submit}}}[G(\tau)]=G(s_{\mathrm{good}})=1,\\
J_G(\pi_{\mathrm{check}})&=\mathbb E_{\pi_{\mathrm{check}}}[G(\tau)]=G(s_{\mathrm{bad}})=0.
\end{aligned}
\label{eq:counterexample-original-returns}
\end{equation}
Thus, $\pi_{\mathrm{submit}}$ uniquely maximizes the original task objective.

Now introduce an action-pattern bonus $b(a)=\lambda\mathbb I\{a=\textsc{check}\}$ with $\lambda>1$, and define $J_{\mathrm{pattern}}(\pi)=\mathbb E_{\pi}[G(\tau)+b(a_0)]$. The modified returns become
\begin{equation}
\begin{aligned}
J_{\mathrm{pattern}}(\pi_{\mathrm{submit}})
&=\mathbb E_{\pi_{\mathrm{submit}}}[G(\tau)+b(a_0)]\\
&=G(s_{\mathrm{good}})+b(\textsc{submit})\\
&=1,\\
J_{\mathrm{pattern}}(\pi_{\mathrm{check}})
&=\mathbb E_{\pi_{\mathrm{check}}}[G(\tau)+b(a_0)]\\
&=G(s_{\mathrm{bad}})+b(\textsc{check})\\
&=\lambda>1.
\end{aligned}
\label{eq:counterexample-modified-returns}
\end{equation}

Therefore, the optimal policy sets differ:
\begin{equation}
\begin{aligned}
\operatorname*{argmax}_{\pi}J_G(\pi)
&=\{\pi_{\mathrm{submit}}\}\\
&\neq\{\pi_{\mathrm{check}}\}\\
&=\operatorname*{argmax}_{\pi}J_{\mathrm{pattern}}(\pi).
\end{aligned}
\label{eq:counterexample-different-optima}
\end{equation}
This establishes Equation~\eqref{eq:action-pattern-noninvariance} and completes the proof of Proposition~\ref{prop:action-pattern-counterexample}.

\subsection{Proof of Theorem~\ref{thm:finite-buffer-decoupling}}
\label{app:proof-finite-buffer-decoupling}

Let $\mathcal D=\{\tau^{(1)},\ldots,\tau^{(N)}\}$ denote the collected trajectories, and let $\mu_{\mathcal D}$ be their empirical distribution. By Definition~\ref{def:finite-buffer-aw}, the advantage-weighted distribution satisfies Equation~\eqref{eq:aw-trajectory-distribution}. Writing $w(\tau)=\exp(A(\tau)/\beta)$, we obtain
\begin{equation}
\begin{aligned}
\mathbb E_{q_{\mathrm{AW}}}[\Psi]
&=\sum_{\tau\in\operatorname{supp}(\mu_{\mathcal D})}q_{\mathrm{AW}}(\tau)\Psi(\tau)\\
&=\sum_{\tau\in\operatorname{supp}(\mu_{\mathcal D})}\frac{\mu_{\mathcal D}(\tau)w(\tau)}{\mathbb E_{\mu_{\mathcal D}}[w]}\Psi(\tau)\\
&=\frac{\mathbb E_{\mu_{\mathcal D}}[w\Psi]}{\mathbb E_{\mu_{\mathcal D}}[w]}.
\end{aligned}
\label{eq:proof-aw-process-expectation}
\end{equation}

Subtracting the original expected process statistic gives
\begin{equation}
\begin{aligned}
\Delta\Psi
&=\mathbb E_{q_{\mathrm{AW}}}[\Psi]-\mathbb E_{\mu_{\mathcal D}}[\Psi]\\
&=\frac{\mathbb E_{\mu_{\mathcal D}}[w\Psi]}{\mathbb E_{\mu_{\mathcal D}}[w]}-\mathbb E_{\mu_{\mathcal D}}[\Psi]\\
&=\frac{\mathbb E_{\mu_{\mathcal D}}[w\Psi]-\mathbb E_{\mu_{\mathcal D}}[w]\mathbb E_{\mu_{\mathcal D}}[\Psi]}{\mathbb E_{\mu_{\mathcal D}}[w]}\\
&=\frac{\operatorname{Cov}_{\mu_{\mathcal D}}(w,\Psi)}{\mathbb E_{\mu_{\mathcal D}}[w]}.
\end{aligned}
\label{eq:proof-process-identity}
\end{equation}

Since $\beta>0$, the exponential weights satisfy $w(\tau)>0$ for every observed trajectory, ensuring $\mathbb E_{\mu_{\mathcal D}}[w]>0$. Therefore, Equation~\eqref{eq:exact-process-shift} shows that $\Delta\Psi$ is nonnegative precisely when $\operatorname{Cov}_{\mu_{\mathcal D}}(w,\Psi)$ is nonnegative. This establishes the process identity and completes the proof of Theorem~\ref{thm:finite-buffer-decoupling}.

\subsection{Proof of Corollary~\ref{cor:monotone-process-improvement}}
\label{app:proof-monotone-process-improvement}

By the law of total covariance, conditioning on the task-centered advantage $A$ gives
\begin{equation}
\begin{aligned}
\operatorname{Cov}_{\mu_{\mathcal D}}(w,\Psi)
&=\operatorname{Cov}_{\mu_{\mathcal D}}\left(\mathbb E_{\mu_{\mathcal D}}[w\mid A],\mathbb E_{\mu_{\mathcal D}}[\Psi\mid A]\right)\\
&\quad+\mathbb E_{\mu_{\mathcal D}}\left[\operatorname{Cov}_{\mu_{\mathcal D}}(w,\Psi\mid A)\right]\\
&=\operatorname{Cov}_{\mu_{\mathcal D}}\left(\exp(A/\beta),h(A)\right),
\end{aligned}
\label{eq:proof-total-covariance}
\end{equation}
where the last equality follows because $w=\exp(A/\beta)$ is deterministic conditional on $A$ and $\mathbb E_{\mu_{\mathcal D}}[\Psi\mid A]=h(A)$.

To establish the sign, let $A'$ be an independent copy of $A$. The covariance admits the representation
\begin{equation}
\begin{aligned}
&\operatorname{Cov}_{\mu_{\mathcal D}}\left(\exp(A/\beta),h(A)\right)\\
&=\frac{1}{2}\mathbb E\left[\left(\exp(A/\beta)-\exp(A'/\beta)\right)\left(h(A)-h(A')\right)\right]\\
&\geq 0,
\end{aligned}
\label{eq:proof-monotone-covariance}
\end{equation}
since both $\exp(\cdot/\beta)$ and $h$ are non-decreasing.

Applying Equation~\eqref{eq:exact-process-shift}, we obtain
\begin{equation}
\Delta\Psi=\frac{\operatorname{Cov}_{\mu_{\mathcal D}}(w,\Psi)}{\mathbb E_{\mu_{\mathcal D}}[w]}\geq 0.
\label{eq:proof-monotone-process-shift}
\end{equation}
This proves Corollary~\ref{cor:monotone-process-improvement}.

\subsection{Proof of Proposition~\ref{prop:return-correlation-not-decisive}}
\label{app:proof-return-correlation-not-decisive}

Consider two tasks $x$ and $y$, each with $n$ trajectories. Let $\mathbf v\in\mathbb R^n$ be a nonconstant centered vector satisfying $\mathbf 1^\top\mathbf v=0$, and let $\mathbf e\in\{0,1\}^n$ indicate the coordinates attaining $\max_i v_i$. Choose $d>0$ and $c>d+\max_i v_i$, and define
\begin{equation}
\begin{aligned}
\mathbf g&=(c\mathbf 1,\ d\mathbf 1+\mathbf v),\\
\boldsymbol\psi&=(\mathbf 0,\ \mathbf e),\\
\mathbf a&=(\mathbf 0,\ \mathbf v).
\end{aligned}
\label{eq:counterexample-return-process}
\end{equation}
The advantage vector $\mathbf a$ follows directly from the task-centered definition in Equation~\eqref{eq:trajectory-advantage}.

Let $\mathbf 1^\top\mathbf e=k$ and $\mathbf e^\top\mathbf v=kv_{\max}$, where $v_{\max}=\max_i v_i>0$. Under the uniform empirical distribution $\mu_{\mathcal D}$,
\begin{equation}
\begin{aligned}
\operatorname{Cov}_{\mu_{\mathcal D}}(G,\Psi)
&=\mathbb E_{\mu_{\mathcal D}}[G\Psi]-\mathbb E_{\mu_{\mathcal D}}[G]\mathbb E_{\mu_{\mathcal D}}[\Psi]\\
&=\frac{\mathbf g^\top\boldsymbol\psi}{2n}-\frac{\mathbf 1^\top\mathbf g}{2n}\frac{\mathbf 1^\top\boldsymbol\psi}{2n}\\
&=\frac{k(d+v_{\max})}{2n}-\frac{c+d}{2}\frac{k}{2n}\\
&=\frac{k(d+2v_{\max}-c)}{4n}.
\end{aligned}
\label{eq:counterexample-return-covariance}
\end{equation}
Choosing $c>d+2v_{\max}$ gives $\operatorname{Cov}_{\mu_{\mathcal D}}(G,\Psi)<0$.

Next, define the elementwise exponential weights $\mathbf w=\exp(\mathbf a/\beta)=(\mathbf 1,\exp(\mathbf v/\beta))$. Writing $z=\exp(v_{\max}/\beta)$ and $S=\mathbf 1^\top\exp(\mathbf v/\beta)$, we obtain
\begin{equation}
\begin{aligned}
\operatorname{Cov}_{\mu_{\mathcal D}}(w,\Psi)
&=\mathbb E_{\mu_{\mathcal D}}[w\Psi]-\mathbb E_{\mu_{\mathcal D}}[w]\mathbb E_{\mu_{\mathcal D}}[\Psi]\\
&=\frac{\mathbf w^\top\boldsymbol\psi}{2n}-\frac{\mathbf 1^\top\mathbf w}{2n}\frac{\mathbf 1^\top\boldsymbol\psi}{2n}\\
&=\frac{kz}{2n}-\frac{n+S}{2n}\frac{k}{2n}\\
&=\frac{k(2nz-n-S)}{4n^2}.
\end{aligned}
\label{eq:counterexample-weight-covariance}
\end{equation}
Since $\mathbf v$ is centered and nonconstant, $v_{\max}>0$, so $z>1$. Moreover, each coordinate satisfies $\exp(v_i/\beta)\leq z$, giving $S\leq nz$. Therefore,
\begin{equation}
\begin{aligned}
2nz-n-S
&\geq 2nz-n-nz\\
&=n(z-1)\\
&>0.
\end{aligned}
\label{eq:counterexample-weight-positivity}
\end{equation}

Combining the two covariance calculations yields
\begin{equation}
\operatorname{Cov}_{\mu_{\mathcal D}}(G,\Psi)<0<\operatorname{Cov}_{\mu_{\mathcal D}}(w,\Psi).
\label{eq:counterexample-opposite-covariances}
\end{equation}
Finally, by Equation~\eqref{eq:exact-process-shift},
\begin{equation}
\begin{aligned}
\Delta\Psi
&=\frac{\operatorname{Cov}_{\mu_{\mathcal D}}(w,\Psi)}{\mathbb E_{\mu_{\mathcal D}}[w]}\\
&=\frac{k(2nz-n-S)}{2n(n+S)}\\
&>0.
\end{aligned}
\label{eq:counterexample-process-improvement}
\end{equation}
This proves Proposition~\ref{prop:return-correlation-not-decisive}.

% BEGIN inlined from sections/additional_process_outcome_figure.tex

% END inlined from sections/additional_process_outcome_figure.tex

\end{document}